# WIFI BASED INDOOR POSITIONING SYSTEM FOR MOBILE ROBOTS BY USING PARTICLE FILTER

Dr. Hikmet Yucel[1]
hikmet.yucel@hyteknik.com

Gulin Elibol[2]
gelibol@ogu.edu.tr

Dr. Ugur Yayan[3]
ugur.yayan@inovasyonmuhendislik.com

November 27, 2020

## ABSTRACT

Mobile robots have capability to work in real-time autonomously. Autonomous behavior is strictly dependent to knowing position of the mobile robot. Positioning of a mobile robot in indoor area is a difficult task for only one sensor information is used. We proposed a system and method to locate mobile robot via fusing signals from WIFI and odometer data via particle filter. In this study, Particle filter is a well-known filter that is used for indoor positioning of mobile robots. The proposed system includes two parts that are RFKON system and evarobot for data collection and experiments. The Received Signal Strength (RSS) measurements of the WiFi access points that are located in the any environment are used to locate a stationary mobile robot in one floor area via SIS Particle Filter. RSS measurements from RFKON database is used and the average location error is 0.7606 and 0.1495 m for 300 and 1000 particles respectively.

*K*eywords  WiFi, indoor positioning, mobile robots, particle filter, localization

## 1. Introduction

Mobile robots (MRs) have capability to work in real-time autonomously. Due to their features the application areas of them are getting expanding from day to day. Recently MRs are used in exploration dangerous or unknown areas [1, 2, 3], transportation in warehouses [4], security environments [5, 6], military applications [7] and education [8]. Localization of MRs (Positioning Systems) is the fundamental parts for all these application areas. The positioning systems are categorized in two groups as indoor and outdoor. Both type of positioning systems use different types of sensors and filters. The data from different type of sensors are fused via filters and the location of mobile robot is obtained and updated simultaneously according to its motion. In indoor positioning systems Particle, Kalman, Extended or Unscented Kalman filters are used to fuse data. Most recent works of particle filter fuses encoder data with ultrasonic [9, 10], RFID [11, 12], camera [13, 14], laser-ranges [15, 16] and WIFI signals [17, 18].

Reference [9] presents a system that locates the mobile robot's position and orientation in indoor area by using ultrasonic sensor and robot's encoder data via Monte Carlo localization method. The real time experiments are done by using differential drive mobile robot and ultrasonic range sensor system that includes four ultrasonic beacons. The aim of this study, analyzing the control parameters of the proposed system. The control parameters are formed from motion model which has uncertainties and noises from ultrasonic range sensors. It is shown that the noise from range sensor measurements and uncertainty of motion model are less than the actual values. However, the best result also has average 0.2210m error and the robot is strayed away from the pre-define path. In [10], localization system is based on particle filter, and ZigBee, ultrasonic sensor and encoder sensors are used. Particle filter is used to reduce error due to transition and rotation. Up to eight ultrasonic sensors are equipped with robot. The real time experiments are taken place 6x9m.

---

[1] Research and Development Department, HY Technic Ltd. Sti., Eskisehir, Turkey
[2] Electrical and Electronics Engineering Dept., Engineering Faculty, Eskisehir Osmangazi University, Eskisehir, Turkey.
[3] Research and Development Department, Inovasyon Muhendislik Ltd. Sti., Eskisehir, Turkey

In [11], RFID tag is used in Particle Filter to locate mobile robot in indoor areas. The proposed algorithm uses robot's position from its observation model and RFID tag to analyses the feasibility of particle filter usage in indoor localization. This algorithm is overcoming simple RFID-based 2D localization problem. The experiments show the proposed system gives better results in dynamic environment. The cost of this system is also high because of RFID tags' expense and also it is impractical for large indoor areas. In [12], particle filter implementation for estimating the pose of tags is proposed by using RFID-equipped robot. The proposed particle filter combines signals from a specially designed RFID antenna system with odometer and a RFID signal propagation model. 6-antenna RFID sensor system provides the robot with a 360-degree view of the tags in its environment. In that experiments, the system estimates the pose of UHF RFID tags in a real-world environment without requiring a priori training or map-building. The system exhibits 0.69m mean range error over robot to tag distances of over 4m in an environment with significant multipath. Evaluation of the RFID system and the particle filter implementation was performed in a 10m × 12m room.

In [13] the localization and navigation of a mobile robot is done via USB webcam and robot's odometer data in office-like places. The webcam is mounted on the robot and used to determine pose of light fixtures. The data from webcam is used with odometer in particle filter to estimate location of the robot. This filter is used to estimate robot's position and orientation in absence of starting location. The real-time experiments are done and the proposed system reduce the cumulative error of sensor. In [14] a depth camera is used to detect ground and the edges of ground. The odometer of robot and these edges are fused via particle filter to locate mobile robot. The depth camera is Kinect and odometer are obtained from robot's encoders. For systems with camera; the cost is high and also has an extra computational cost due to processing of images from vision sensors.

Laser rangefinder is fused with odometer by Particle Filter in [15] and 240-degree laser range finder is used. The results are proving that fusing data from laser and odometer increases the accuracy of estimated position. In [16], localization system uses odometer from gyroscope and a 3D laser scanner. The 3D laser scanner has uniform measuring-point distribution. Localization is implemented on the sensor platform using a particle filter on a 2D grid map generated by projecting 3D points on to the ground. The sensor platform with a particle filter fuses the gyro-assisted odometer and the 3D laser scanner. The localization system uses 20 grid maps which are generated by projecting 3D points. The 20 map contains rich information from the wide-view 3D scanner while it has a small data structure.

WIFI is another sensor to use in particle filter for indoor positioning. In [17], Machine Learning (ML) algorithms are used to find position of mobile robot. Beside that Monte Carlo Localization (MCL) algorithm is used to improve the previous results by taking into consideration of the encoder data. The proposed system uses 5000 particles and has 0.27 m mean error. In [18], the localization system uses the odometer data and particle filter with RSS measurements. A map is constructed and that map includes 106 unique access points and 500 particles is used in particle filter. The real-time experiments are done in one floor (818 m path) and the localization mean error is 0.7 m.

In literature, there aren't many studies that are focused on fusing WIFI and odometer data for mobile robot localization in indoor areas. We proposed a system to locate mobile robot via fusing signals from WIFI and odometer data via particle filter. The proposed system includes two parts that are RSS database and HASKON. The radio-frequency map of RSS from WIFI nodes is constructed in first part. Five WIFI nodes are used to collect RSS and construct radio-frequency map. Locating the mobile robot via using particle filter is performed in HASKON. Radio-frequency map and odometer are used in particle filter in HASKON for mobile robot localization.

The rest of this paper is organized as follows. The proposed sensor fusion algorithm is explained in Section II. Section III describes RFKON system and evarobot. The experimental results of the proposed system are given in Section IV. Conclusion and future works are given in the final section.

## 2. PROPOSED SENSOR FUSION ALGORITHM

In our proposed algorithm, position of the mobile robot is acquired from particle filter. Particles are located randomly in indoor area. Each particle pretends like a mobile robot. In prediction step, odometer data is used and particles are moved according to mobile robot's position. The RSS measurements of the particles are obtained from the radio map database. The RSS measurement of each particle is obtained from the closest landmark's measurement. In correction step, weights of the particles are calculated by likelihood function. The estimated position of the mobile robot is acquired from average of the particles after resampling. The flow chart of the working principle of the proposed system is shown in Figure 1.

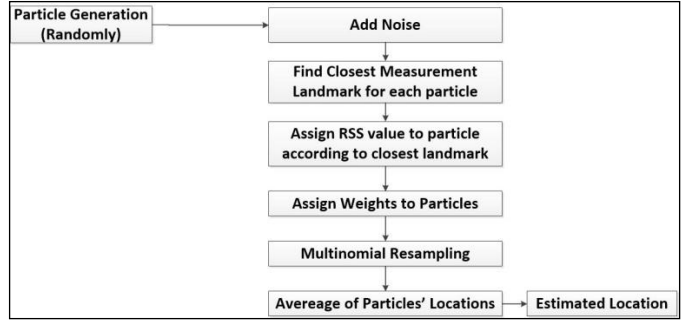

Figure 1: Flow chart of the Particle Filter

Particle filter is a Sequential Monte Carlo Estimation [19]. The working principle of this filter depends on the particles that are distributed to space. All particles have posterior probabilities that are named as weights. These weights are updating via the new observations. In addition to that, the particles can be resampled according to their weights. The weighted mean of the particles is the output of the filter. Particle filters have many variations such that Sequential Importance Sampling, Sampling Importance Resampling, Rejection, Auxiliary etc. In this study Sequential Importance Sampling Particle Filter (SIS) is used.

The first step of the proposed algorithm generating particles. The particles are generated in space according to (1) for generic PF.

$$p(x_n|y_n) \approx \frac{1}{N_p} \sum_{n=1}^{N_p} \delta(x_n - x_n^i) \equiv \hat{p}(x_n|y_n) \quad (1)$$

where $x_n^i$ are the particles that are generating from conditional probability of $x_n$ and $y_n$. This conditional probability means that when the event of $y_n$ is happened, what is the probability of occurrence of $x_n$. In terms of particle filter, this probability shows the possibility of where the mobile robot's location is.

In SIS filter particles are sampled according to (2).

$$x_n^i \sim q(x_n|x_{0:n-1}^i, y_{0:n}) \quad x_{0:n}^i = \{x_{0:n-1}^i, x_n^i\} \; for \; i = 1, \ldots, N_p \quad (2)$$

In this study particles are generating randomly in 10x10 area where the experiments are performed. Each particle is considered as 2D point, thus each one has 2 coordinate information. $N_p$ particles are sampled such as

$$x_n^i = [Area\;Length * Random\;Number_i \quad Area\;Width * Random\;Number_i\;] \quad (3)$$

The generated 300 particles according to indoor area are shown in Figure 2. The stars are the landmarks of the RF map. The circles are the particles.

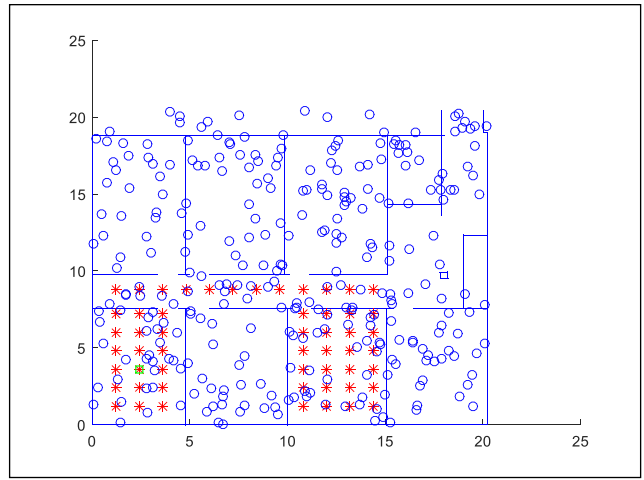

Figure 2: Generated particles and landmark in indoor area

The particles have weights. These weights are calculated for every sampled particle. The weight calculation for generic particle filter is shown in (4).

$$W_n(x_n) = \frac{p(y_n|x_n)p(x_n)}{q(x_n|y_n)} \quad (4)$$

In SIS filter, the importance weights are calculated such as

$$W_n^i = W_{n-1}^i \frac{p(y_n|x_n^i)p(x_n^i|x_{n-1}^i)}{q(x_n^i|x_{0:n-1}^i, y_n)} \quad (5)$$

Weights are determined by using likelihood ($p(y_n|x_n)$), knowledge of the model ($p(x_n|y_n)$) and transition density ($p(x_n|x_{n-1})$).

Each particle has own RSS measurement. These measurements are obtained from the database. The RSS measurement of landmarks is assigned to a particle according to the distance between particle and the landmark. Assuming position of the robot is $(X_1, Y_1)$ and measurement landmarks is $(X_i, Y_i)$.

$$Closest\ Point = \arg\min(\sqrt{((X_1 - X_i)^2 + (Y_1 - Y_i)^2)}) \quad (6)$$

The importance weights are calculated via Likelihood function for this study. The likelihood function is shown in (7).

$$W_n^i = \frac{1}{\left(\sigma\sqrt{2\pi}\right)^4 e^{\frac{-(x-\hat{x})^2}{2\sigma^2}}} \quad (7)$$

The calculated importance weights are in a large range and they should be normalized. The normalization ($\widetilde{W}_n(x_n^i)$) is important for resampling. The normalized weights are calculated according to (8).

$$\widetilde{W}_n(x_n^i) = \frac{W_n(x_n^i)}{\sum_{j=1}^{N_p} W_n(x_n^j)} \quad (8)$$

The resampling condition must be controlled to perform resampling. This condition is calculating efficient particles and is shown in (9).

$$\frac{1}{\sum_{i=1}^{N_p}(\widetilde{W}_n^i)^{\wedge}2} > \frac{N_p}{2} \quad (9)$$

If the given condition is not satisfied, then resampling must be done and a new particle set must be generated. The resampling strategy is based on the particles' importance weights. The particles that have high importance weights are selected and the rest of the particles are eliminating. The selected particles are sampled again and a new particle set is generated. The Multinomial Resampling method is used in this study. Summary of this resampling method is shown below:

- Generate Np random numbers with uniform distribution via weights
- Extract new particle set from generated Np numbers via selecting the particles
- Duplicate the particles according to their weights
- Assign Weights to new particle set: Weights=1/Np.

The estimation of position is the mean value of resampled particles. The estimation is calculated such that;

$$P_{MR} = \frac{1}{Np}\sum_{i=1}^{N_p} x_n^i \quad (10)$$

## 3.   SYSTEM OVERVIEW AND RADIO FREQUENCY MAP

RFKON is a system that is constructed to solve indoor positioning problem for mobile devices and robots. It contains three units named as GEZKON, HASKON, and KONSENS. GEZKON is a mobile application which is responsible for collecting RSS values and self-localization of mobile devices. HASKON is a special hardware and software for self-localization of mobile robots. KONSENS is a server that collects RSS values from the access points (APs) in the region to construct / update / calibrate the RFKON database [20]. This database is then used by GEZKON and HASKON for self-positioning by applying machine learning, positioning and fusion algorithms. KONSENS is also used to estimate position of access points in the region. Sensor nodes (SensDug) that are working under KONSENS, emit RF signals for positioning and collect RF signals in the environment. Sensor nodes are also responsible for communication within all units. This communication is established by DDS layer. This indoor localization scenario is given in Figure 3.

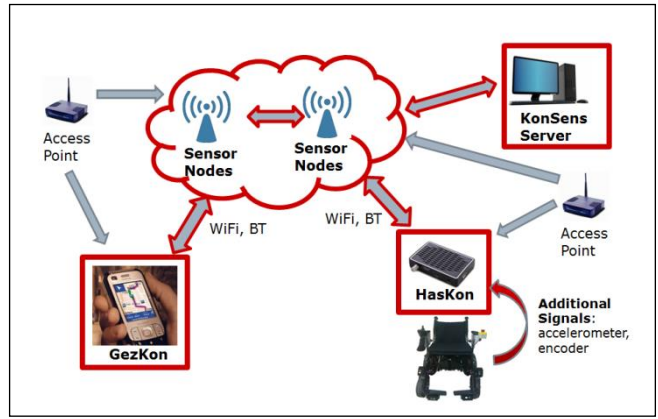

Figure 3: Indoor location scenario

The hardware of the KONSENS Server is shown in Figure 4. Besides this hardware, SensDug is used for network communication for roaming.

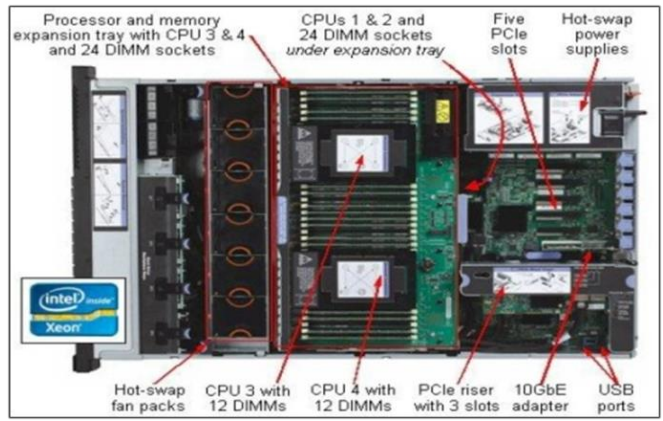

Figure 4: Konsens server hardware

KONSENS runs on Ubuntu 14.04 LTS Server operating system. It contains four nodes named as KONSENS Control, SensDug DDS, GEZKON Web, and HASKON DDS. All these nodes use DDS layer as a middle layer for establishing data transferring process between KONSENS and other units (HASKON, and GEZKON). These nodes are represented in Figure 5.

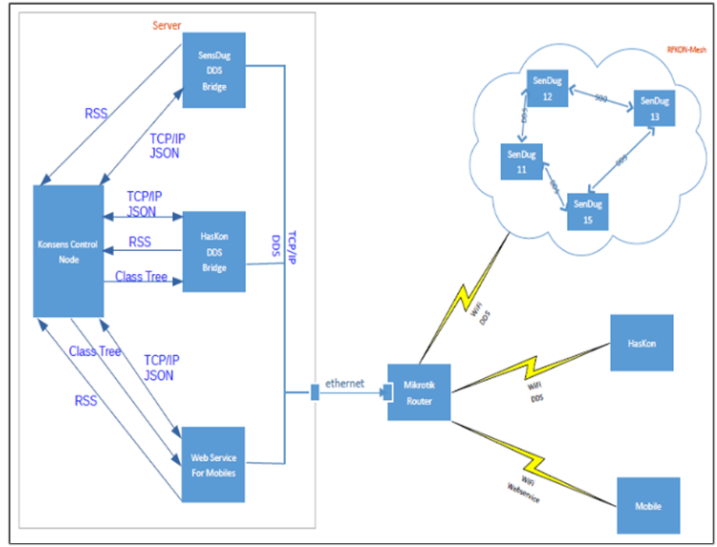

Figure 5: Konsens system overview

KONSENS control node is deal with database construction phase. It collects RSS values via SensDug DDS, HASKON DDS node and GEZKON Web node writes the collected values into MongoDB database. Because of environmental dynamics (wall, furniture, people...) and device diversity, the RF database must be calibrated/updated for accurate positioning. SensDug collects RSS values from the APs in the region and sends these values to KONSENS control node using TCP/IP protocol. Data are in a JSON data format with MAC addresses and corresponding RSS values. Gezkon Web node collects RSS values from the APs in the region and sends these data to KONSENS control node for constructing RF database named as RFKON_MB_WiFi and RFKON_MB_BT according to sensor type. Data sending process is achieved using TCP/IP protocol.

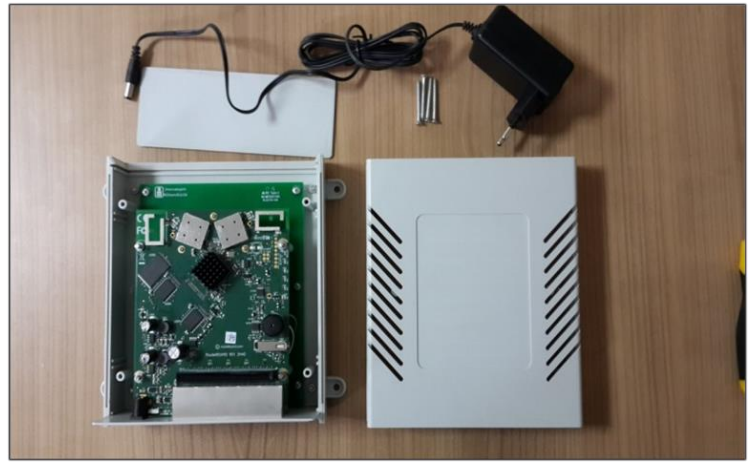

Figure 6: Hardware architecture of Konsens Sensor Network

The hardware of SensDug contains Mikrotik RB951Ui and Raspberry Pi B+ that is illustrated in Figure 6. They are chosen because of low price, ease of use, and ease of supply. It also involves USB WiFi and BT dongles for collecting/emitting WiFi and BT RSS values.

Haskon is tested on evarobot platform. Evarobot (http://wiki.ros.org/Robots/evarobot) has open source software, ROS compatible electronics, modular design for sensory equipment, and different kind of simple and complex sensors for various tasks. Evarobot's modular mechanical design (see Fig. 7) consist of two parts body and the upper platform. Body contains motors, encoders, wheels, battery components and basic level sensors like infrared and sonar. Upper platform contains advanced sensors like RGB-D Camera, 3600 lidar, and head angle reference system. Also, there is GAZEBO model and plugins of evarobot for conducting experiments in simulation environment.

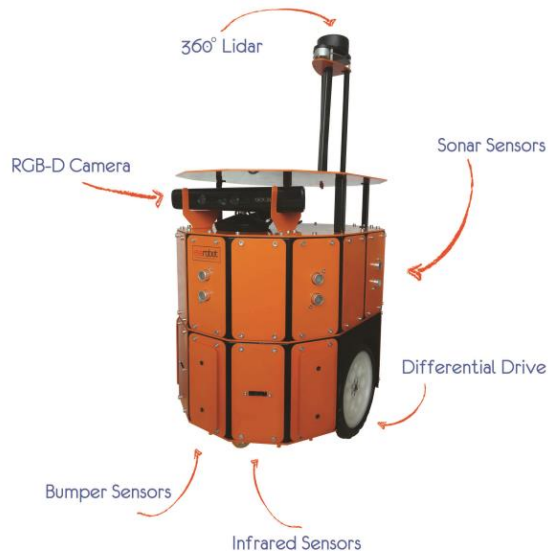

Figure 7: Evarobot modular mechanical design and equipped sensors

Evarobot is managed by Electronic Control Unit (EKB) consist of two parts which are Low Level Control (DSK) and High-Level Control (YSK). EKB-DSK is built on Raspberry Pi 2. EKB-DSK is responsible for gathering data from encoder, sonar, infrared and battery sensors, controlling the motor speed. Also, EKB-DSK is responsible from security software when any emergency situation (if robot is too close to any obstacle, then do not apply any command from the EKB-YSK i.e.) occurs. EKB-YSK is motherboard of a miniPC or any laptop and it is responsible for gathering data from laser, kinect, camera and communicate with other robot or server via Wi-Fi. EKB-YSK also implements a high-level control that enables the behave of the robot autonomously. EKB structure and default equipped sensory for evarobot is can be seen in Figure 8.

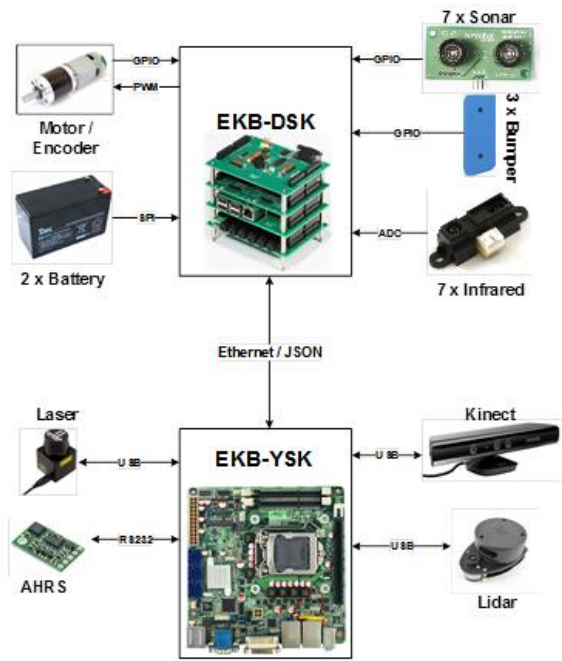

Figure 8: Default Evarobot units

## 4. EXPERIMENTAL RESULTS

In this section experiments are explained and discussed. The particle filter is implemented in MATLAB according to Section III. The mobile robot is at one of the measurement landmarks where RSS are measured before and saved in a database. The mobile robot's initial position is unknown and the aim is finding its location via particle filter. Firstly, particles are generated randomly at indoor area. Their weights are set to be same value that is 1/Np at the beginning.

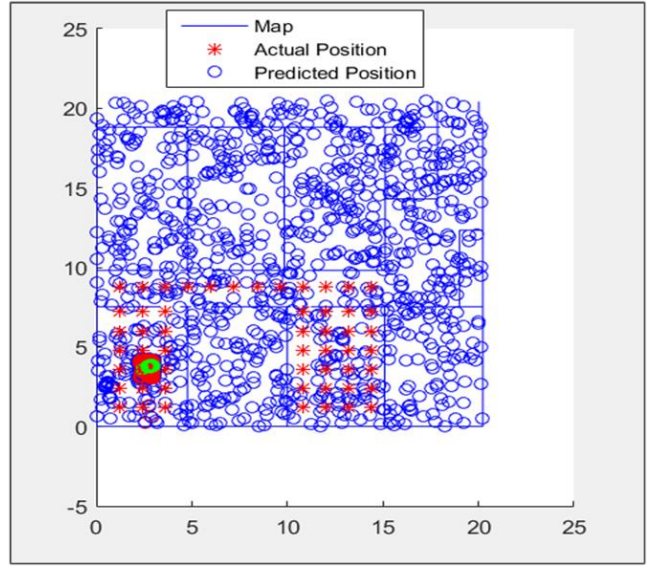

Figure 9: Generated particles on the map of floor

For each mobile robot's movement, that means each encoder data is obtained, location of all particles is updated according to displacement of robot. The fundamental idea is based on robot's motion, but in these experiments mobile robot does not move and only stands. Although the mobile robot is stationary, to be more practical, noise is added to particles' locations. The added noise is a random number between [-0.05 0.05] m uniform distribution.

The RSS measurements are obtained from RFKON database and data from only 5 sensor nodes are used. The RSS measurements of particles (predicted measurements) are acquired by calculating the Euclidean distance between particle and the actual measurement locations (measurement landmarks). After finding the closest distance between particle and actual measurement landmark, the RSS measurement of this closest measurement landmark is assigned to particle. Figure 10 shows the distances between particles and actual measurement landmarks.

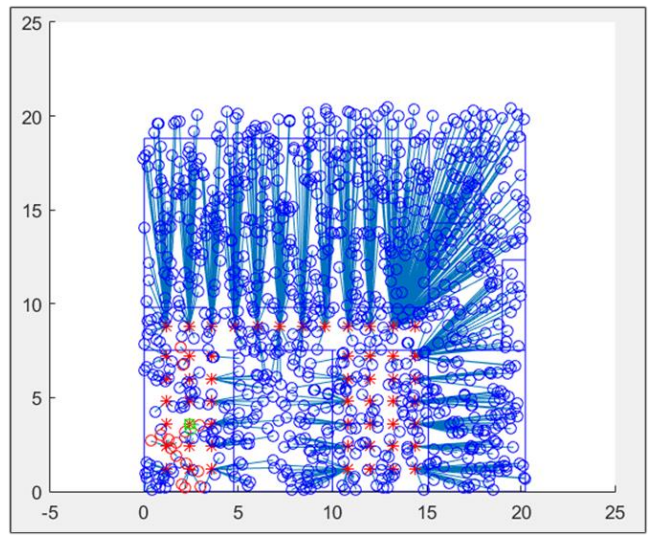

Figure 10: Closest measurement landmark for each particle

In first iteration (particles are not resampled yet), locations of the particles are in a large range. However, the outermost particle is also assigned an actual measurement landmark. After particles' measurements (RSS) are found, weights of the particles are calculated via using updated locations of the particles and the RSS measurements. Weights are calculated via likelihood function in (6) and normalized. The difference between particles' RSS measurement and actual measurements are used as "$\hat{x}$" and "x".

After weights are calculated, the particles are resampled via using these values in multinomial resampling method. With this method, particles are selected, duplicated and after all Np particles are obtained. These resampled particles have higher chance to be the mobile robot's location. Figure 11 shows the locations of the particles after resampling.

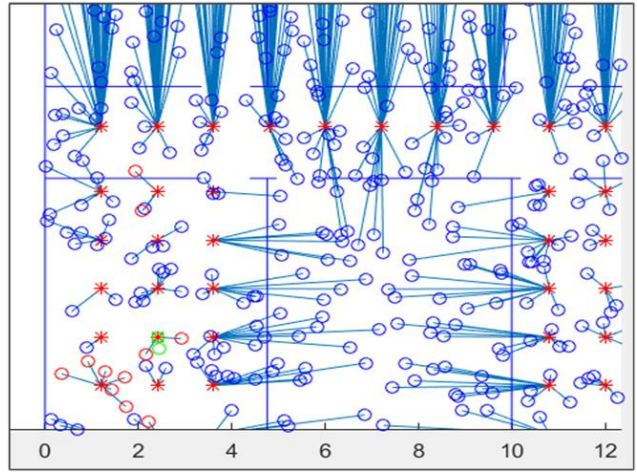

Figure 11. Resampled particles and closest measurement landmarks

After many iterations, resampled partciles are concentrated upon a few locations. An illustration is shown in Figure 12 and more closely version is shown in Figure 13.

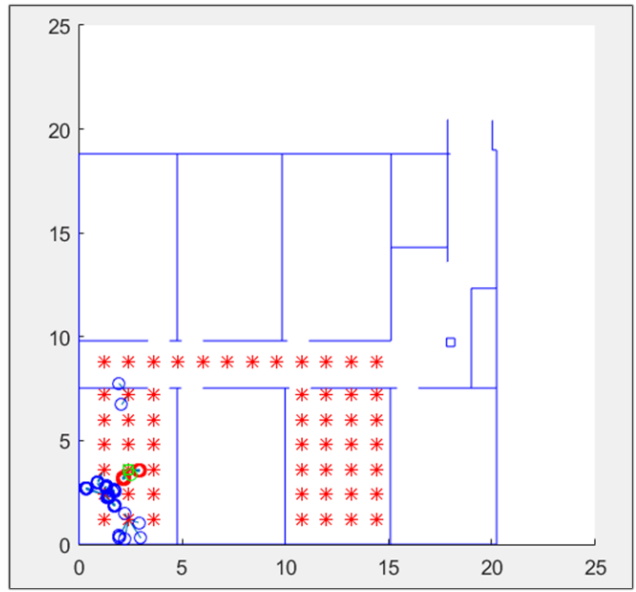

Figure 12: The stationary mobile robot position and particles

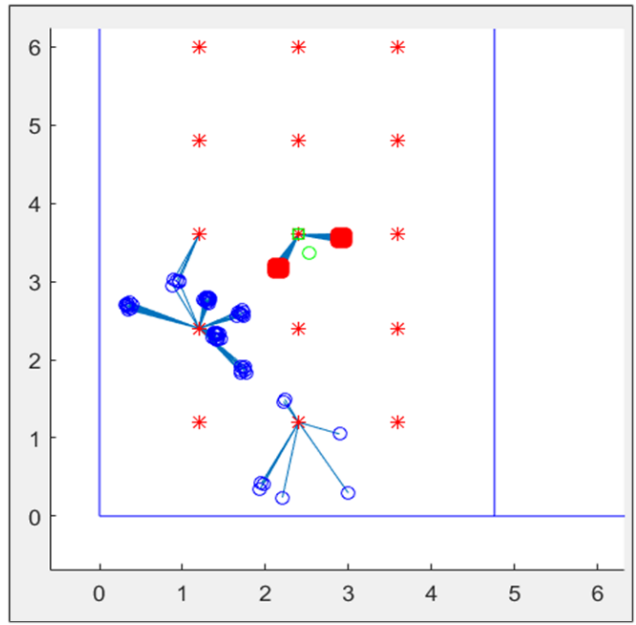

Figure 13: The stationary mobile robot position and particles (clearer)

The blue circles are the particles before resampling and the red circles are the resampled particles. The mean of locations of the resampled particles are calculates and assigned as estimated position. Green circles are estimated positions of the mobile robot and green square is the initial unknown mobile robot position.

When 300 and 1000 particles are used the estimated position is shown in Figure 14 and 15 respectively.

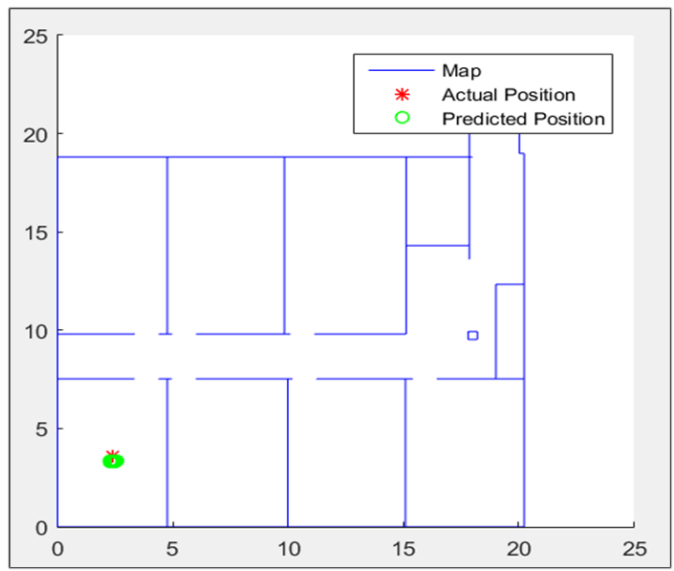

Figure 14: Estimated location and the mobile robot's actual position for 300 particles

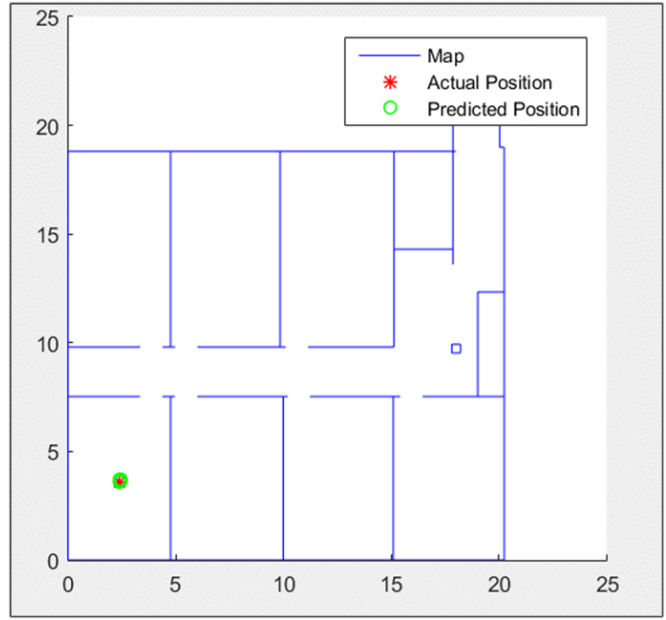

Figure 15: Estimated location and the mobile robot's actual position for 1000 particle

The initial position of the mobile robot is at (2.4, 3.6) m. The estimated position for 300 and 1000 particles is shown in Table 1. Due to the random particle generation, the particle filter is testes 10 times with 300 and 1000 particles. The location error is calculated and the average error is 0.7606m and 0.1445m for 300 and 1000 particles respectively.

Table 1: Estimated positions and errors

| $N_p$ | 300 Estimated Position | Error (m) | 1000 Estimated Position | Error (m) |
|---|---|---|---|---|
| 1 | (2.110, 3.3734) | 0.368 | (2.3924, 3.6787) | 0.0791 |
| 2 | (0.4525, 2.24) | 2.3754 | (2.2438, 3.7596) | 0.2233 |
| 3 | (2.1819, 3.818) | 0.3084 | (2.4608, 3.5035) | 0.1141 |
| 4 | (2.1197, 3.278) | 0.4271 | (2.6503, 3.7645) | 0.2995 |
| 5 | (1.93, 3.125) | 0.6682 | (2.4952, 3.4588) | 0.1703 |
| 6 | (2.421, 3.5995) | 0.0210 | (2.5102, 3.6299) | 0.1142 |
| 7 | (2.631, 3.5496) | 0.2361 | (2.5517, 3.5786) | 0.1532 |
| 8 | (2.726, 0.8694) | 2.7499 | (2.5787, 3.5801) | 0.1798 |
| 9 | (2.261, 3.5752) | 0.1411 | (2.4846, 3.6878) | 0.1219 |
| 10 | (2.101, 3.5157) | 0.3107 | (2.4278, 3.5724) | 0.0392 |

## 5. Conclusion

In conclusion, this study proposes a method for positioning of mobile robots that is focused on fusing WIFI and odometer data for mobile robot localization in indoor areas by using Particle Filter. The proposed system includes two parts that are RFKON system and evarobot hardware for data collection and experiments. To demonstrate the performance of the positioning systems, various tests were conducted experimental results gave us the average positioning error is 0.7606 and 0.1495 m for 300 and 1000 particles respectively.

Future work includes real environment tests, trajectory tracking experiments and cross track error analysis for proposed method and system.

**Acknowledgements**: This work is supported by The Scientific and Technological Research Council of Turkey (TUBITAK) under "RF based indoor positioning system (RFKON)" grant number 1130024.